\documentclass[letterpaper]{article} %

\usepackage{aaai23}  %
\usepackage{times}  %
\usepackage{helvet}  %
\usepackage{courier}  %
\usepackage[hyphens]{url}  %
\usepackage{graphicx} %
\usepackage{natbib}  %
\usepackage{caption} %
\usepackage{algorithm}
\usepackage{algorithmic}

\usepackage{booktabs}           %
\usepackage{multirow}           %
\usepackage{amsfonts}           %
\usepackage{subfig}
\usepackage{amsmath}

\usepackage{newfloat}
\usepackage{listings}
\DeclareCaptionStyle{ruled}{labelfont=normalfont,labelsep=colon,strut=off} %
\floatstyle{ruled}
\newfloat{listing}{tb}{lst}{}
\floatname{listing}{Listing}
\title{Event Detection on Dynamic Graphs\footnote{Long version of \textit{Graph Macro Dynamics with Self-Attention for Event Detection} published in DLG-AAAI'23.}}
\author{
    Mert Kosan\textsuperscript{\rm 1},
    Arlei Silva\textsuperscript{\rm 2},
    Sourav Medya\textsuperscript{\rm 3},
    Brian Uzzi\textsuperscript{\rm 4},
    Ambuj Singh\textsuperscript{\rm 1}
}
\affiliations{
    \textsuperscript{\rm 1}University of California, Santa Barbara\\
    \textsuperscript{\rm 2}Rice University\\
    \textsuperscript{\rm 3}University of Illinois Chicago\\
    \textsuperscript{\rm 4}Northwestern University\\
    \{mertkosan, ambuj\}@ucsb.edu, arlei@rice.edu, medya@uic.edu, uzzi@kellogg.northwestern.edu
}

\usepackage{soul}
\usepackage{url}
\usepackage{epstopdf}
\usepackage{float}
\usepackage[utf8]{inputenc}
\usepackage{graphicx}
\usepackage{amssymb}
\usepackage{booktabs}
\usepackage{algorithm}
\usepackage{comment}
\usepackage{booktabs} %
\usepackage{graphicx,array}
\usepackage{balance}
\usepackage{color,soul,xspace}

\usepackage{comment}
\usepackage{epsfig}
\usepackage{subfig}
\usepackage{mathrsfs,mdwlist,enumitem,amsthm}

\def\dyged{DyGED\xspace}

\newtheorem{defn}{\textbf{Definition}}

\begin{document}

\maketitle

\begin{abstract}
Event detection is a critical task for timely decision-making in graph analytics applications. Despite the recent progress towards deep learning on graphs, event detection on dynamic graphs presents particular challenges to existing architectures. Real-life events are often associated with sudden deviations of the normal behavior of the graph. However, existing approaches for dynamic node embedding are unable to capture the graph-level dynamics related to events. In this paper, we propose \dyged, a simple yet novel deep learning model for event detection on dynamic graphs. \dyged learns correlations between the graph macro dynamics---i.e. a sequence of graph-level representations---and labeled events. Moreover, our approach combines structural and temporal self-attention mechanisms to account for application-specific node and time importances effectively. Our experimental evaluation, using a representative set of datasets, demonstrates that \dyged outperforms competing solutions in terms of event detection accuracy by up to 8.5\% while  being more scalable than the top alternatives. We also present case studies illustrating key features of our model.
\end{abstract}

\section{Introduction}
\label{sec::intro}

Event detection on dynamic graphs is a relevant task for effective decision-making in many organizations \cite{li2017detecting,leetaru2013gdelt}. In graphs, entities and their interactions are represented as (possibly attributed) nodes and edges, respectively. The graph dynamics, which changes the interactions and attributes over time, can be represented as a sequence of snapshots. Events, identified as snapshot labels, are associated with a short-lived deviation from normal behavior in the graph.

As an example, consider the communication inside an organization, such as instant messages and phone calls \cite{romeroWWW16}. Can the evolution of communication patterns reveal the rise of important events---e.g., a crisis, project deadline---within the organization? While one would expect the content of these communications to be useful for event detection, this data is highly sensitive and often private. Instead, can events be discovered based only on structural information (i.e. message participants and their attributes)? For example, Romero et al. \cite{romeroWWW16} have shown that stock price shocks induce changes (e.g., higher clustering) in the structure of a hedge fund communication network. Shortly after the 2011 earthquake and tsunami, Japanese speakers expanded their network of communication on Twitter \cite{lu2014network}. 

Given the recent success of deep learning on graphs \cite{kipf2016semi,hamilton2017inductive,wu2019comprehensive,georgousis2021graph} in node/graph classification, link prediction, and other tasks, it is natural to ask whether the same can also be useful for event detection. In particular, such an approach can combine techniques for graph classification \cite{zhang2018end,errica2020fair} and dynamic representation learning on graphs \cite{nicolicioiu2019recurrent,seo2018structured,pareja2019evolvegcn,zhao2019t}. However, a key design question in this setting is whether to detect events based on the micro (node) or macro (graph) level dynamics. More specifically, the micro dynamics is captured via the application of a pooling operator to dynamic node embeddings \cite{nicolicioiu2019recurrent,pareja2019evolvegcn}. For the macro dynamics, static snapshot embeddings are computed via pooling and their evolution is modeled via a recurrent architecture (e.g. an LSTM) \cite{seo2018structured,zhao2019t}. Each of these approaches has implicit assumptions about the nature of events in the data.

\begin{figure}
\centering
\subfloat[Generic architecture for event detection \label{fig::event_detection_illustrative_generic}]{
\includegraphics[keepaspectratio, width=0.42\textwidth]{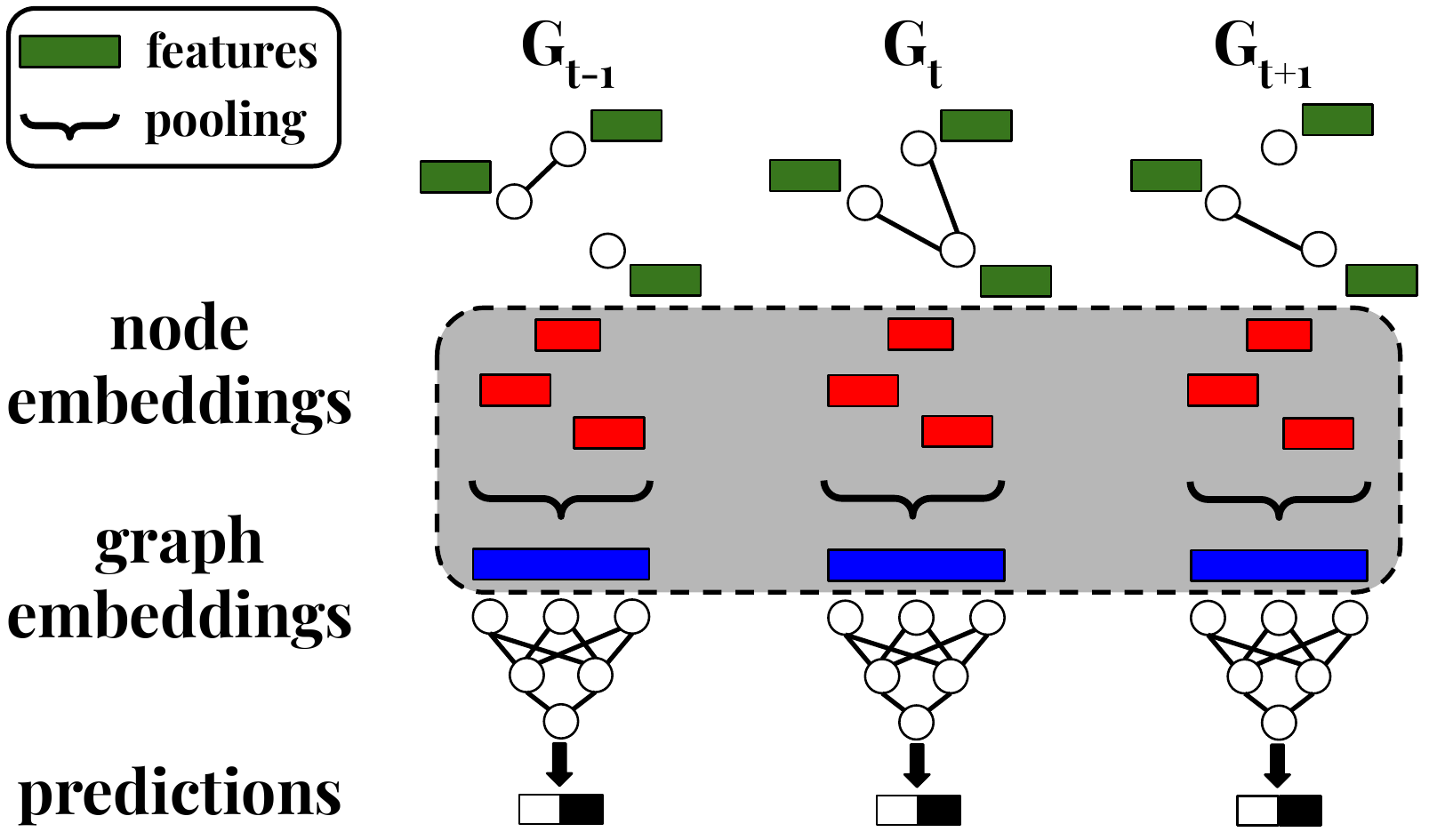}
}

\subfloat[Learning graph embeddings based on micro dynamics]{
\includegraphics[keepaspectratio, width=0.42\textwidth]{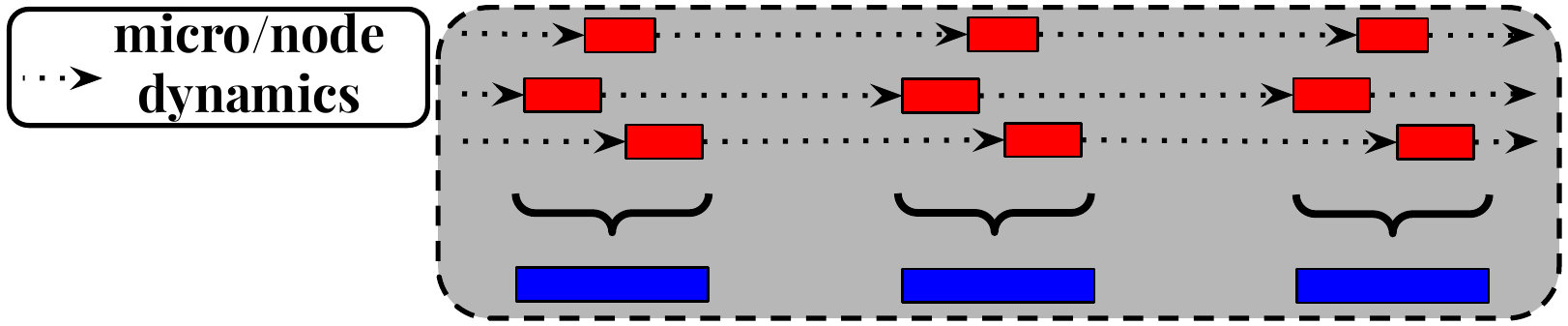}
}

\subfloat[Learning graph embeddings based on macro dynamics]{
\includegraphics[keepaspectratio, width=0.42\textwidth]{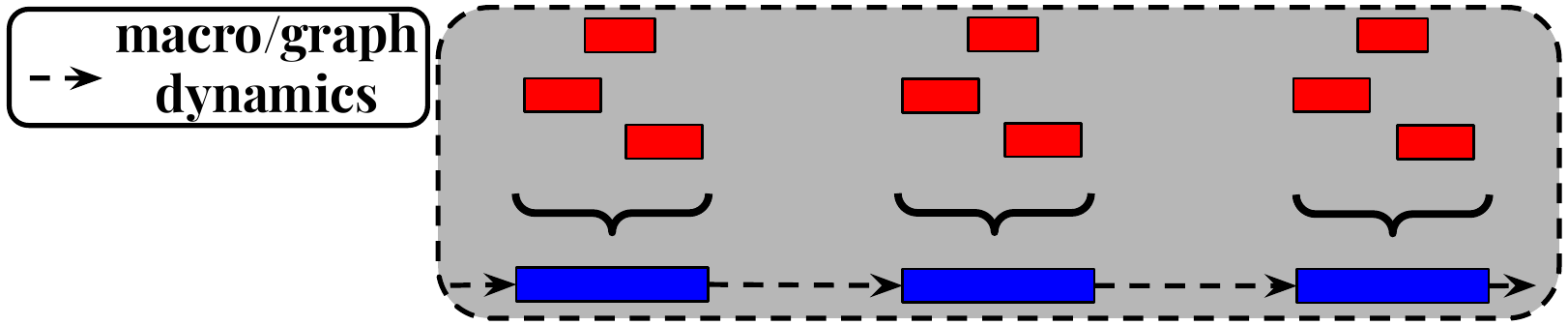}
}
\caption{(a) Event detection on dynamic graphs based on a generic deep learning architecture. (b) At the micro scale, the dynamics is captured at node level using a temporal GNN architecture and then pooled for graph-level classification. (c) At the macro scale, the dynamics is captured at the graph level using an RNN over pooled (static) GNN node embeddings. Our work investigates how the dynamics at different scales affects event detection performance.  \label{fig::event_detection_illustrative}}
\end{figure}

Figure \ref{fig::event_detection_illustrative} shows two event detection architectures, one based on micro and another based on macro dynamics. While they both apply a generic architecture shown in Figure \ref{fig::event_detection_illustrative_generic}, they differ in the way dynamic representations for each graph snapshot are generated.

To illustrate the difference between micro and macro dynamics, let us revisit our organization example. For simplicity, we will assume that the pooling operator is the average. Dynamic node embeddings are learned (non-linear) functions of the evolution of an employee's attributed neighborhood. These local embeddings are expected to be revealing of an employee's communication over time. Thus, (pooled) micro embeddings will capture average dynamic communication patterns within the organization. On the other hand, by pooling static node embeddings, we learn macro representations for the communication inside the organization at each timestamp. The recurrent architecture will then capture dynamic communication patterns at the organization level. Pooling and the RNN thus act as (spatial/temporal) functions that can be composed in different ways---e.g. f(g(x)) vs g(f(x))---each encoding specific inductive biases for event detection. We will show that the choice between micro and macro models has significant implications for event detection performance.

This paper investigates the event detection problem on dynamic graphs. We propose \dyged (Dynamic Graph Event Detection), a graph neural network for event detection. \dyged combines a macro model with structural and temporal self-attention to account for application-specific node and time importances. To the best of our knowledge, our work is the first to apply either macro dynamics or self-attention for the event detection task. Despite its simplicity, differing from more recent approaches based on micro dynamics, \dyged outperforms state-of-the-art solutions in three representative datasets. These findings also have implications for other graph-level analytics tasks on dynamic graphs, such as anomaly detection, regression, and prediction.

One of the strengths of our study is its extensive experimental evaluation. While the event detection problem has been studied by a recent paper \cite{deng2019learning}---based on micro dynamics---our work provides key insights into some of the challenges and possible strategies for effective event detection. This is partly due to our representative list of datasets covering mobility, communication, and user-generated content data. Moreover, we present a few case studies illustrating the key features of our approach. Our main contributions are as follows:

\begin{itemize}
    \item We present the first study comparing micro and macro deep learning architectures for event detection on dynamic graphs, showing the importance of this design choice for event detection performance;\\
    
    \item We propose \dyged, a simple yet novel deep learning architecture for event detection based on macro dynamics. \dyged applies both structural and temporal self-attention to enable the effective learning of node and time dependent weights;\\
    
    \item We compare \dyged against several baselines---mostly based on a micro model---using three datasets. Our results show that \dyged outperforms the baselines by up to 8.5\% while being scalable. We also provide case studies illustrating relevant features of our model (e.g. how its embeddings can be used for diagnosis).
\end{itemize}

\section{Related Work}
\label{sec::related_work}
\textbf{Event detection on graphs:} There is a diverse body of work on event detection using graphs and other types of structured data \cite{sakaki2010earthquake,ramakrishnan2014beating,atefeh2015survey} in the literature. Moreover, other popular tasks such as anomaly \cite{ranshous2015anomaly,rayana2015less}, change point \cite{akoglu2010event,peel2015detecting}, and intrusion \cite{staniford1996grids} detection are related to unsupervised event detection \cite{chen2014non,rayana2015less,rozenshtein2014event}. Here, we focus on a supervised version of the problem where graph snapshots are labeled depending on whether an event happened within the snapshot's time window. We assume that events are defined based on an external source of information, not being fully identifiable from---but still being correlated with---the observed data \cite{li2017detecting,deng2019learning}. Some studies distinguish the event detection from the event forecasting problem, as the second also allows for predictions of events far into the future \cite{ramakrishnan2014beating}. However, such a distinction is less relevant when events are external to the data \cite{deng2019learning}. Recently in \cite{hu2021time}, the authors apply a dynamic graph to identify events on time series. We focus on the detection of events with graphs as inputs.

The classical approach for event detection, including in the case of graphs, is to rely on features designed by experts \cite{aggarwal2012event,ning2018staple}. A framework for automatically combining multiple social network metrics, such as modularity and clustering coefficient, as features for terrorist activity detection using a neural network is introduced in \cite{li2017detecting}. However, these metrics, which are based on expert knowledge, may not generalize to many applications, and they are potentially non-exhaustive---i.e. other relevant metrics might be missing.

\textbf{Graph kernels:} In machine learning, kernel methods, such as Support Vector Machines, have motivated a long literature on graph kernels \cite{kriege2020survey}. A kernel is a function that computes the similarity between two objects in a particular application and thus their design often requires expert-knowledge. In the case of graph kernels, graphs are compared based on their substructures---e.g., node neighborhoods \cite{kriege2012subgraph}, graphlets \cite{shervashidze2009efficient}, random walks \cite{kashima2003marginalized}. However, recent studies have shown that such features are often outperformed by those learned directly from data \cite{zhang2018end}.

\textbf{Temporal node embeddings:} Modeling temporal evolution of nodes based on their connections over time is a well-studied problem in the literature. The temporal information can be encoded using temporal point process \cite{trivedi2019dyrep}, random walks on temporal edge orderings \cite{nguyen2018continuous} or a joint optimization of temporal node embeddings \cite{singer2019node}. While these methods focus on micro embeddings, \cite{lu2019temporal} proposes taking into account macro information by incorporating the number of modified edges in the graph. However, we notice that \cite{lu2019temporal} does not apply to our setting since it does not find node embeddings for every timestamp.

\textbf{Graph neural networks (GNNs) for dynamic graphs:}
Deep learning on graphs is an effort to reproduce the success achieved by deep neural networks (e.g. CNNs, RNNs) on graphs \cite{gori2005new,duvenaud2015convolutional}. A key advance was the introduction of Graph Convolutional Networks (GCNs) \cite{kipf2016semi}, which outperform traditional approaches for semi-supervised learning. Later, GraphSage \cite{hamilton2017inductive} and Graph Attention Networks (GATs) \cite{velivckovic2017graph} were proposed to increase the scalability and exploit the attention mechanism in GNNs, respectively.

Following the work listed above, there has been an outburst of extensions of GNNs for dynamic graphs \cite{skarding2020foundations}. One approach is to apply standard GNNs to a static graph where multiple snapshots are represented as a single multi-layered graph \cite{yan2018spatial,guo2019attention} with additional temporal edges. More recently, several studies have combined recursive architectures, such as LSTMs, with GNNs. This can be achieved via macro models, which stack graph-level GNN representations as a sequential process \cite{taheri2019learning,seo2018structured}. Another alternative are micro models, which apply the sequential process at the node level to generate dynamic node embeddings \cite{sankar2020dysat,pareja2019evolvegcn,guo2019attention}. 

Dynamic GNNs have been applied mostly for local tasks (e.g. node classification). On the other hand, the most popular graph-level task for GNNs is graph classification \cite{niepert2016learning,lee2018graph,ying2018hierarchical}, which assumes the input to be static. Our paper is focused on event detection on dynamic graphs, a challenging graph-level task also studied in \cite{deng2019learning}---one of our baselines. Different from their work, \dyged applies self-attention at node and time domains to capture the macro dynamics correlated with events. Self-attention was also recently applied in \cite{li2019semi} and \cite{sankar2020dysat}, which were focused on graph classification and link prediction, respectively. We show that \dyged equipped with self-attention outperforms state-of-the-art event detection methods.

\section{Problem Definition}
\label{sec::problem_definition}

Supervised event detection on dynamic graphs consists of learning how to detect events based on a few recent graph snapshots using training data (i.e. labeled events). 

\begin{defn}
\textbf{Dynamic Graph:} A dynamic graph $\mathbb{G}$ is a sequence of \hspace{.02cm} $T$ discrete snapshots $\langle G_1,G_2,\dots,G_T \rangle$ where $G_t$ denotes the graph at timestamp $t$. $G_t$ is a tuple $(V, E_t, W_t, X_t)$ where $V$ is a fixed set of $n$ vertices, $E_t$ is a set of $m_t$ undirected edges, $W_t:E_t\rightarrow \mathbb{R}_+$ are edge weights, and $X_t:V\rightarrow \mathbb{R}^d$ gives $d$ features for each node. 
\end{defn}

In our earlier example regarding an organization (Section \ref{sec::intro}), nodes in $V$ represent members of the organization. An edge is created in $E_t$ whenever their associated members exchange a message during a time interval $t$ and weights $W_t$ might be numbers of messages exchanged. Finally, features $X_t$ might include an individual's job position (static) and the total number of messages received by them during the time interval $t$ (dynamic).

\begin{defn}
\textbf{Event Label Function:} We define the function $\ell(G_{t-k:t})$ $\in \{0,1\}$ to be an event labelling function of order $k$, where $G_{t-k:t} = \langle G_{t-k},$ $G_{t-k+1}, \ldots, G_{t} \rangle$, and such that:

\begin{equation}
    \nonumber
    \ell(G_{t-k:t}) =
    \begin{cases}
        1, &  \text{if an event occurs at time $t$}\\
        0, & \text{otherwise}
    \end{cases}
\end{equation}

\label{def::event_label}
\end{defn}

Events might depend not only on the current graph snapshot $G_t$ but also on the $k$ previous snapshots $G_{t-k}, \ldots, G_{t-1}$. This allows the function $\ell$ to model events that depend on how the graph changes. One can define a similar function $\ell_{\Delta}$ for the early detection (or forecasting) of events $\Delta$ snapshots into the future.

\begin{defn}
\textbf{Event Detection Problem:} Given a set of training instances $\mathbb{D}$, composed of pairs $(G_{t-k:t},\ell(G_{t-k:t}))$, learn a function $\hat{\ell}$ that approximates the true $\ell$ for unseen snapshots.
\end{defn}

We treat event detection as a classification problem with two classes. To evaluate the quality of the learned function $\hat{\ell}$ we apply a traditional evaluation metric (AUC \cite{fawcett2006introduction}) from supervised learning. In this paper, we propose $\hat{\ell}$ to be a neural network.

\section{Proposed Model: DyGED}
\label{sec::methods}

We describe \dyged (Dynamic Graph Event Detection), a simple yet novel deep learning architecture for event detection on dynamic graphs. \dyged combines a Graph Convolutional Network and a Recurrent Neural Network to learn the macro (i.e. graph-level) dynamics correlated with labeled events. This backbone architecture is further enhanced by self-attention mechanisms in the structural and temporal domains. First, we introduce the main components of our architecture. Next, we describe \dyged and some of its variations.

We introduce notations for a few basic operations to describe our architectures in a compact form. A column-wise concatenation $[M_1, \ldots, M_t]: \mathbb{R}^{n\times m_1} \times \ldots \times \mathbb{R}^{n\times m_t} \to \mathbb{R}^{n\times (m_1+\ldots+m_t)}$ maps a sequence of matrices $M_1, \ldots, M_t$ to a new matrix $M$ such that $(M_t)_{i,j}\!=\!M_{i,\sum_{r=1}^{t-1}m_r+j}$. Similarly, we denote a row-wise concatenation as $[M_1; \ldots; M_t] = [M_1^{\intercal}, \ldots, M_t^{\intercal}]^{\intercal}$.

\subsection{Main Components}

We describe the main components of our neural network architecture for event detection on dynamic graphs (\dyged).

\subsubsection{Graph Convolutional Network:}

GCNs are neural network architectures that support the learning of $h$-dimensional functions $\mathbf{GCN}(A,X): \mathbb{R}^{n\times n} \times \mathbb{R}^{n \times d} \to \mathbb{R}^{n \times h}$ over vertices based on the graph adjacency matrix $A$ and features $X$. For instance, a 2-layer $GCN$ can be defined as follows:

\begin{equation}
\nonumber
\mathbf{GCN}(A,X)=\sigma \left(\hat{A} \ \sigma \left(\hat{A}XW^{(0)}\right)\ W^{(1)} \right)
\label{eqn::gcn}
\end{equation}

\vspace{2mm}
where $\hat{A}=\!\tilde{D}^{-\frac{1}{2}}\tilde{A}\tilde{D}^{-\frac{1}{2}}$ is the normalized adjacency matrix with $\tilde{D}$ as weighted degree matrix and $\tilde{A}\!=\!I_n\!+\!A$ with $I_n$ being an $n\times n$ identity matrix. $W^{(i)}$ is a weight matrix for the $i$-th layer to be learned during training, with $W^{(1)} \in \mathbb{R}^{d\times h'}$, $W^{(2)} \in \mathbb{R}^{h'\times h}$, and $h$ ($h'$) being the embedding size in the output (hidden) layer. Moreover, $\sigma(.)$ is a non-linear activation function such as $\mathbf{ReLU}$.

\subsubsection{Pooling}
\label{sec::pooling}
The output of the GCN described in the previous section is an embedding matrix $Z_t$ for each graph snapshot $G_t$. In order to produce an embedding $\mathbf{z}_t$ for the entire snapshot, we apply a pooling operator $\text{v-Att}(Z_t): \mathbb{R}^{n\times h}\to \mathbb{R}^h$. In particular, our model applies the self-attention graph pooling operator proposed in \cite{li2019semi}:

\begin{equation}
    \nonumber
\mathbf{z_t}=\mathbf{\text{v-Att}}(Z_t)=\mathbf{softmax}(\mathbf{w.tanh}(\Phi Z_t^T))Z_t
\end{equation}

\vspace{3mm}
where $\Phi\in\mathbb{R}^{h\times h}$ and $w\in\mathbb{R}^h$ are learned attention weights.

Intuitively, v-Att re-weights the node embeddings enabling some nodes to play a larger role in the detection of events. In our experiments, we will show that these attention weights can be used to identify the most important nodes for our task.

\subsubsection{Recurrent Neural Network}

We assume that events are correlated with the graph (i.e. macro) dynamics. Thus, our model applies an RNN to learn dynamic graph representations. More specifically, we give the pooled snapshot embeddings $\mathbf{z}_t$ as input to a standard Long Short-Term Memory $\mathbf{LSTM}(\mathbf{z_{t-1}'}, \mathbf{z_t}):\mathbb{R}^h\times \mathbb{R}^h\to \mathbb{R}^h$ to produce dynamic graph representations:

\begin{equation}
\nonumber
\mathbf{z_t'} = \mathbf{LSTM}(\mathbf{z_{t-1}'}, \mathbf{z_t})    
\end{equation}

Notice that $\mathbf{z_t'}$ is based on a sequence of static embeddings, instead of each node's (micro) dynamics. In our experiments, we will compare these two approaches using a diverse collection of datasets. 

\subsubsection{Temporal Self-Attention}

The RNN component described in the previous section enables our architecture to capture the graph dynamics via embeddings $\mathbf{z_t'}$. However, complex events might not be correlated only with the current graph representation but a window $Z'_t = [\mathbf{z'_{t-k}}; \ldots; \mathbf{z'_t}]$. For instance, in mobility-related events (e.g. sports games), changes in the mobility dynamics will arise a few hours before the event takes place. Moreover, these correlations might vary within a dataset due to the characteristics of each type of event. Thus, we propose a self-attention operator $\mathbf{\text{t-Att}}(Z'_t):\mathbb{R}^{(k+1)\times h}\to \mathbb{R}^h$ for aggregating multiple dynamic embeddings:

\begin{equation}
\nonumber
\mathbf{z''_t}=\mathbf{\text{t-Att}}(Z'_t)=\mathbf{softmax}(\mathbf{w'.tanh}(\Phi' Z_t^{'T}))Z'_t
\end{equation}

\vspace{4mm}
where $\Phi' \in \mathbb{R}^{h\times h}$ and $w' \in \mathbb{R}^h$ are learned attention weights.

Similar to v-Att (Section \ref{sec::pooling}), t-Att enables the adaptive aggregation of dynamic embeddings. To the best of our knowledge, we are the first to apply a similar self-attention mechanism---which might be of independent interest---in dynamic GNN architectures.

\subsubsection{Classifier and Loss Function}
\label{sec::classifier_loss}

The final component our model is a Multi-Layer Perceptron $MLP(z_t''):\mathbb{R}^h\to \mathbb{R}^2$ that returns (nonlinear) scores for each possible outcome (i.e., event/no event) $Y_t$. Given training data with event labels $\ell(G_{t-k:t})$, the parameters of our model are learned (end-to-end) by minimizing the \textit{cross-entropy} of event predictions $Y=\{Y_{k+1}, \ldots, Y_T\}$. Note that event detection is a highly imbalanced problem---i.e. events are rare. We address this challenge by weighting our loss function terms with class ratios \cite{8943952}. As a result, false negatives are more  penalized than false positives.

\begin{equation}
\nonumber
     -\sum_{t=k+1}^{T}(1-x)\ell(G_{t-k:t})\log{(Y_{t,1})}+x(1-\ell(G_{t-k:t}))\log{(Y_{t,2})}
 \label{eqn::event_detection_loss}
\end{equation}

\vspace{3mm}
where $\ell$ is the event label from Definition \ref{def::event_label}. Moreover, $x$ and $1 - x$ positive (i.e., events) and negative sample ratios in the training set, respectively.

\subsection{\dyged and its Variants}

\begin{algorithm}[t]
	\caption {\dyged Forward Algorithm}
	\label{alg:dyged}
	 {\small
	\begin{algorithmic}[1]
	 \REQUIRE Sequence of snapshots $G_{t-k:t}$, previous dynamic state $z_{t-k-1}'$
	 \ENSURE Event probability
	 \FOR{$\tau \in \{t-k, \ldots, t\}$}
	 \STATE $Z_{\tau} \leftarrow GCN(G_{\tau}, X_{\tau})$
	 \STATE $z_{\tau} \leftarrow \text{v-Att}(Z_{\tau})$
	 \STATE $z_{\tau}' \leftarrow LSTM(z'_{\tau-1},z_{\tau})$
	 \ENDFOR
	 \STATE $z_t'' \leftarrow \text{t-Att}([z'_{t-k}; \ldots; z'_t])$
	 \STATE \textbf{return} $MLP(z_t'')$
   \end{algorithmic}}
\end{algorithm}

Algorithm \ref{alg:dyged} provides an overview of the forward steps of \dyged. It receives a sequence of snapshots $G_{t-k:t}$ and the previous dynamic (LSTM) state $z_{t-k-1}'$ as inputs. The output is the event probability for $G_t$. Notice that our assumption that macro dynamics of the graph is correlated with events of interest leads to a simple and modular model. Steps 3 and 5 correspond to the structural and temporal self-attention, respectively. In order to evaluate some of the key decisions involved in the design of \dyged, we also consider the following variations of our model: \\

\begin{itemize}
    \item \dyged-CT (with contatenation): Replaces the LSTM (step 4) and t-Att (step 5) operators by a concatenation, with $z_t''=([z_{t-k}, \ldots, z_t])$. \\
    \item \dyged-NL (no LSTM): Removes the LSTM operator (step 4) from Algorithm \ref{alg:dyged}, with $z_t'' = \text{t-Att}([z_{t-k}; \ldots; z_t])$.\\
    \item \dyged-NA (no attention): Removes the temporal self-attention operator t-Att (step 5), with $z_t''= z_t'$.\\ %
\end{itemize}

\subsubsection{Time Complexity}\label{sec:time_complexity} Table \ref{tab:time_complexity} shows the time complexities for different variations of \dyged discussed in this section. Our methods are scalable as the time complexities are linear with the number of vertices ($n$) and edges ($m$) in the input graph.

\begin{table}[htbp]
\centering
\setlength{\tabcolsep}{4pt}
\renewcommand{\arraystretch}{1}
\scriptsize
\begin{tabular}{ccccc}%
\toprule
& \textbf{GCN + Pooling} & \textbf{LSTM} & \textbf{t-Att} & \textbf{MLP} \\
\midrule
\textit{\dyged-CT} & $O((mh+nh^2)l_1)$ & - & - & $O(kh^2 + h^2l_2)$ \\
\textit{\dyged-NL} & $O((mh+nh^2)l_1)$ & - & $O(kh^2)$ & $O(h^2l_2)$ \\
\textit{\dyged-NA}& $O((mh+nh^2)l_1)$ & $O(h^2)$ & - & $O(h^2l_2)$ \\
\textit{\dyged}& $O((mh+nh^2)l_1)$ & $O(h^2)$ & $O(kh^2)$ & $O(h^2l_2)$ \\
\bottomrule
\end{tabular}
\caption{Time complexities of our methods: $m, n, h, l_1, k, l_2$ denote numbers of edges, nodes, embedding dimension, layers in GCN, past snapshots, layers in MLP. We assume the initial node feature dimension for GCN, input, cell, and output dimension for LSTM equal to embedding dimension $h$.
\label{tab:time_complexity}}
\end{table}

\section{Experiments}
\label{sec::experiments}

We evaluate \dyged---which is our approach for event detection on dynamic graphs---and its variations using a diverse set of datasets. We compare our solutions against state-of-the-art baselines for event detection, graph classification, and dynamic GNNs in terms of accuracy (Sec. \ref{sec::results_accuracy}) and efficiency (Sec. \ref{sec::dyged_efficiency}). We also provide a visualization of prediction scores to give more insight into the results (Sec. \ref{sec::visualization_event_score}. Furthermore, we present more studies to have a more detailed picture of the effectiveness of \dyged across representative application scenarios using event embeddings (Sec. \ref{sec::event_diagnosis}), importance via attention (Sec. \ref{sec::attention_results}), and ablation study including feature and pooling operator variations (Sec. \ref{sec::ablation_study}). The implementation of DyGED and datasets are available online\footnote{\url{https://www.github.com/mertkosan/DyGED}}.

\begin{table}[ht]
\footnotesize
\centering
\setlength{\tabcolsep}{3pt}
\renewcommand{\arraystretch}{1}
\begin{tabular}{ccccc}%
\toprule
& \textbf{NYC Cab} & \textbf{Hedge Fund} & \textbf{TW} & \textbf{TW-Large} \\
\midrule
\textit{\#Nodes (avg)}& 263 & 330 & 300 & 1000 \\
\textit{\#Edges (avg)}& 3717 & 557 & 1142 & 10312 \\
\textit{\#static features}& 6 & 5 & 300 & 300 \\
\textit{\#dynamic features}& 3 & 3 & 3 & 3 \\
\textit{\#Snapshots}& 4464 & 690 & 2557 & 2557 \\
\textit{Snap. Period}& hour & day & day & day \\
\textit{\#Events}& 162 & 55 & 287 & 287 \\
\bottomrule
\end{tabular}
\caption{The statistics of the datasets. TW is Twitter Weather.}
\label{tab::dataset_stats}
\end{table}

\begin{table*}[ht]
\centering
\begin{tabular}{p{10em}ccccc}%
\toprule
& \textbf{Method} & \textbf{{NYC Cab}}& \textbf{{Hedge Fund}} & \textbf{{TW}} & \textbf{{TW Large}} \\
\midrule
\multirow{3}{10em}{\textbf{{Baselines: \\ Micro Dynamics}}}
& EvolveGCN & ${0.842}^{**} \pm 0.008$ & ${0.718}^{**} \pm 0.011$ & ${0.782}^{**} \pm 0.012$ & ${0.731}^{**} \pm 0.011$ \\
& ASTGCN  & ${0.903}^{**} \pm 0.003$ & ${0.753}^{*} \pm 0.022$ & ${0.747}^{**} \pm 0.018$ & ${0.722}^{**} \pm 0.014$ \\
& DynGCN  & ${0.901}^{**} \pm 0.003$ & ${0.679}^{**} \pm 0.030$ & ${0.713}^{**} \pm 0.012$ & ${0.709}^{**} \pm 0.007$ \\
\midrule
\textbf{{Classification}} & DiffPool  & ${0.887}^{**} \pm 0.003$ & ${0.690}^{**} \pm 0.020$ & ${0.766}^{**} \pm 0.014$ & ${0.728}^{**} \pm 0.007$ \\
\midrule
\multirow{4}{10em}{\textbf{{Proposed: \\ Macro Dynamics}}} 
& \textit{\dyged-CT} & $\underline{0.910} \pm 0.009$ & ${0.776} \pm 0.012$ & ${0.775}^{**} \pm 0.014$ & ${0.743}^{**} \pm 0.014$ \\
& \textit{\dyged-NL} & $\textbf{0.912} \pm 0.004$ & $0.779^{*} \pm 0.012$ & ${0.791}^{**} \pm 0.014$ & $\underline{0.752}^{*} \pm 0.020$ \\
& \textit{\dyged-NA} & ${0.896}^{**} \pm 0.004$ & $\underline{0.784}^{*} \pm 0.014$ & $\underline{0.800}^{*} \pm 0.009$ & ${0.734}^{**} \pm 0.012$ \\
& \textit{\dyged} & ${0.905}^{*} \pm 0.004$ & $\textbf{0.787} \pm 0.015$ & $\textbf{0.810} \pm 0.012$ & $\textbf{0.760} \pm 0.014$ \\
\bottomrule
\end{tabular}
\caption{AUC scores of event detection methods. The highest and second highest values for each column are in bold and underlined, respectively. Our methods, accounting for macro dynamics, achieve the best results, outperforming the best baseline (ASTGCN) by 4.5\% on average and up to 8.5\% (on Twitter Weather). We performed a paired t-test comparing the best model against the others (markers ** and * indicate p-value $<$ .01 and $<$ .05, respectively). %
 \label{tab:quality_methods}}
 \end{table*}
 
\subsection{Datasets}
\label{sec::data}

Table \ref{tab::dataset_stats} shows the main statistics of our datasets. The snapshot period is the interval $[time_t, time_t + \Delta{p})$ covered by each snapshot $G_t$, where $\Delta{p}$ denotes the period. These datasets are representatives of relevant event detection applications. NYC Cab is an example of a mobility network with geo-tagged mass-gathering events (e.g., concerts, protests). Hedge Fund is a communication network for decision making in high-risk environments---as in other business settings and emergency response. Twitter Weather relates user-generated content with extreme events (e.g., terrorist attacks, earthquakes).

\textbf{NYC Cab:} Dataset based on sports events and hourly numbers of passengers transported between cab zones in NYC.\footnote{\url{https://www1.nyc.gov/site/tlc/about/tlc-trip-record-data.page}} Nodes, edges, and their weights represent cab zones, inter-zone trips, and numbers of passengers transferred, respectively. Static node features are lat-long coordinates, boroughs, lengths, areas, and service zones. Dynamic features are the node degree, betweenness centrality, and clustering coefficient within a snapshot (i.e. one hour period). Baseball games involving the Yankees or the Mets in NYC are the events of interest.

\textbf{Hedge Fund \cite{romeroWWW16}:}
Dynamic network of employees at a hedge fund and their communications. Stock market shocks between January 2009 and September 2011 are the events. Nodes, edges, and edge weights represent employees, communications, and the total number of messages exchanged, respectively. Node features are employee's personal information such as company name, hiring time, gender, and position. We consider the node degree, betweenness centrality, and clustering coefficient as dynamic features. Each snapshot covers activities in a day, and events are price shocks---unexpected changes  \cite{romeroWWW16}--- in the S\&P500. 

\textbf{Twitter Weather:}
Dataset integrating weather-related tweets and significant weather events in the US from 2012 to 2018. Tweets were extracted from a large corpus made available by the Internet Archive\footnote{\url{https://archive.org/details/twitterstream}}. Nodes, edges, and edge weights represent English words, word co-occurrences, and the number of co-occurrences, respectively. Starting from a small set of weather-related words, we employed an existing algorithm \cite{wang2016identifying} to expand the set to $300$ words. Pre-trained word2vec\footnote{\url{https://code.google.com/archive/p/word2vec/}} \cite{mikolov2013efficient} vectors were applied as static node features, whereas dynamic features are the node degree, betweenness centrality, and clustering coefficient during a one day interval. Weather events---with monetary damage of at least \$50M---were collected from the US National Climatic Data Center records.\footnote{\url{https://www.ncdc.noaa.gov/stormevents/}} We also created a larger version of Twitter Weather with 1000 words.

\subsection{Experimental Settings}

\subsubsection{Baselines:} We consider recent approaches that either focus on micro (node-level) dynamics \cite{deng2019learning,pareja2019evolvegcn,guo2019attention} or are designed for graph classification \cite{ying2018hierarchical}. If it is necessary, we apply our v-Att module (see Section \ref{sec::pooling}) to get graph embeddings from node embeddings.
\begin{itemize}
\item \textbf{DynGCN \cite{deng2019learning}:} State-of-the-art architecture for event detection that combines representations from a GCN at each snapshot with a temporal encoder that carries information from the past.
\item \textbf{EvolveGCN \cite{pareja2019evolvegcn}:} Combines recurrent and graph convolutional neural networks to generate dynamic node embeddings.
\item \textbf{ASTGCN \cite{guo2019attention}: }Graph convolutional  network originally proposed for traffic prediction. It combines spatial and temporal-attention mechanisms. We adapt ASTGCN to our problem setting by considering $k$ previous time dependencies instead of daily, weekly, and monthly ones.
\item \textbf{DiffPool \cite{ying2018hierarchical}: }Computes graph embeddings via a differentiable graph pooling method. Because this model is designed for classification, it does not account for the dynamics.
\end{itemize}

\subsubsection{Other Settings}

\textbf{Train/Test Splits:} We evaluate the methods using $p$-fold nested cross-validation \cite{bergmeir2012use}, where $p$ was set based on event frequency. Each method runs 20 times per train/test split (at least 100 repetitions/method), and we report average results. Train/test splits of $3720/744$, $575/115$, and $2192/365$ snapshots are applied for the NYC Cab, Hedge Fund, Twitter Weather datasets, respectively.

\textbf{Ratio of positive and negative samples: }Overall percentages of positive samples (events) are 3.62\%, 7.97\%, and 11.22\% for the NYC Cab, Hedge Fund, and Twitter Weather dataset, respectively. As we use nested cross-validation, train/test event ratios vary (depending on the fold) from/to 3.46/4.18\%-3.61/3.37\%, 3.57/12.5\%-8.37/7.14\% and 11.17/18.23\%-13.53/4.14\% for NYC Cab, Hedge Fund, and Twitter Weather respectively.

\textbf{Hyper-parameters:} We tune the hyper-parameters of our methods and baselines with a grid search. We find that training using Adam optimization with learning rate, dropout rate, and the batch size set to $0.005$, $0.2$, and $100$, respectively, works well for our methods. The number of GCN layers, MLP layers, the size of the embedding set to $2$, $2$, and $64$, respectively, are good choices for model parameters.

\textbf{Quality metric:} We compare the quality of the predictions by the methods using the Area under the ROC curve (AUC) \cite{fawcett2006introduction}.

\textbf{Hardware:} We run our experiments on a machine with NVIDIA GeForce RTX 2080 GPU (8GB of RAM) and 32 Intel Xeon CPUs (2.10GHz and 128GB of RAM).

\subsection{Event Detection Accuracy\label{sec::results_accuracy}}

Table \ref{tab:quality_methods} shows the event detection accuracy results in terms of AUC. For the approaches that consider a sliding window, reported results are the best ones among window sizes ($k+1$) varying from one to five. The optimal window size for all these methods is either four or five. We use only static node features in our main experiments, and the effect of dynamic features will be shown in Section \ref{sec::ablation_study}.

Results show that \dyged outperforms the competing approaches in all datasets. In particular, \dyged outperforms ASTGCN---best baseline---by 0.2\%, 4.5\%, 8.5\%, and 5.2\% for NYC Cab, Hedge Fund, Twitter Weather, and Twitter Weather Large, respectively (4.5\% on average). Notice that, different from most baselines, our approach captures the macro-dynamics correlated with events. \dyged-NL and \dyged, which adopt temporal self-attention, achieve the best results indicating that it enables the learning of adaptive weights for different snapshots/times. 
Moreover, \dyged-NA and \dyged---using recurrent neural network to capture the macro dynamics---achieve better performance for the Hedge Fund and Twitter Weather datasets.

\begin{figure*}[htbp]
\centering
\includegraphics[width=0.99\linewidth]{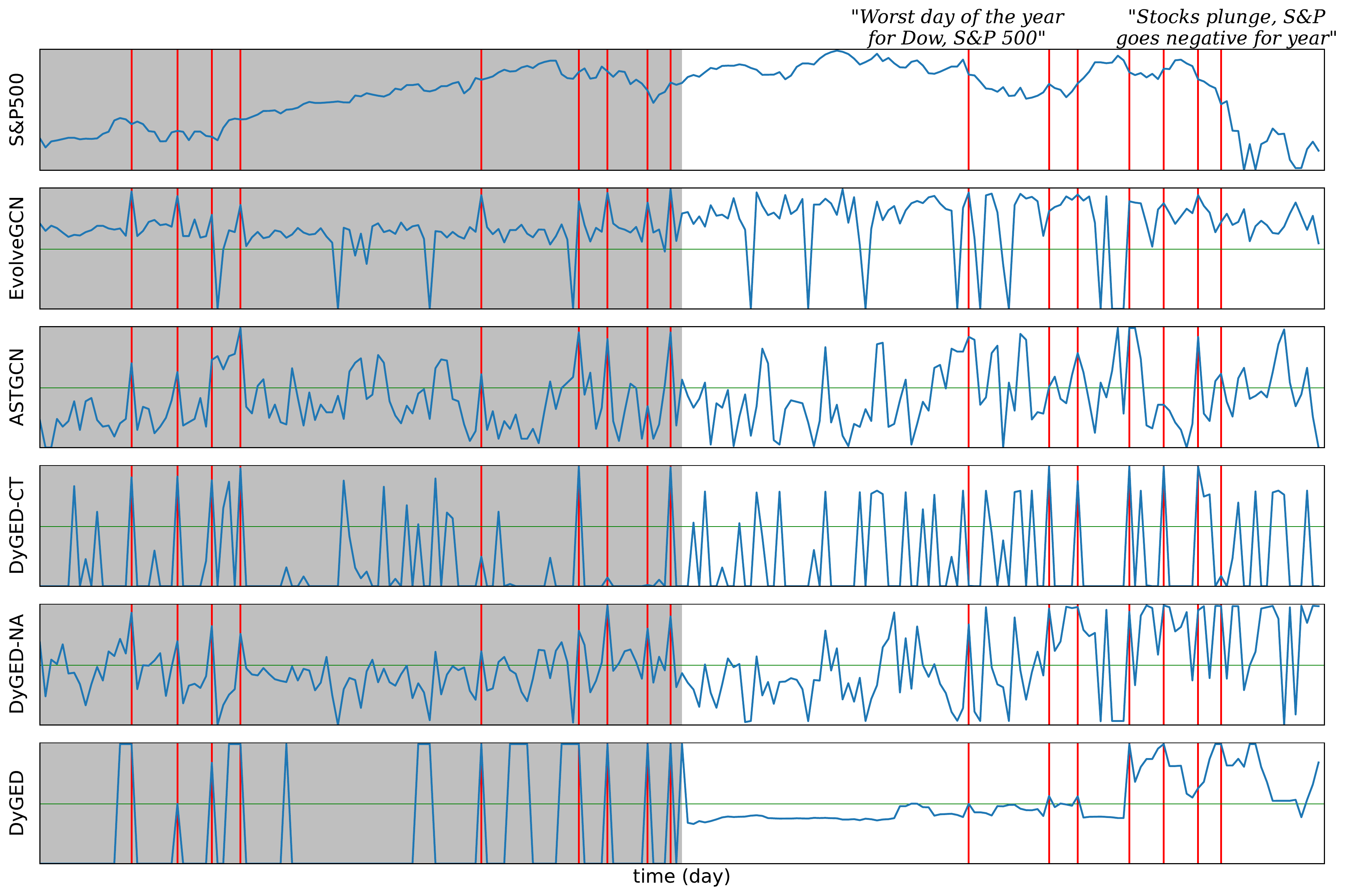}
\caption{Scaled prediction scores ($[0,1]$) given by some of the event detection methods for Hedge Fund. Events are marked as vertical (red) lines. The first (grey) half covers some of the training events, while the remaining (white) is in the test set. Some events are annotated for illustration. Hedge Fund annotations are headlines from CNBC on the day of the event. Our method \dyged can identify several of the events while keeping the false positive rate lower. \label{fig::visualization}}
\end{figure*}

\subsection{Visualization of Event Scores}
\label{sec::visualization_event_score}

Besides evaluating event detection approaches in terms of accuracy, it is also important to analyze how event scores are related to true events in the data. This is particularly relevant when events are determined based on a threshold.

Figure \ref{fig::visualization} shows the prediction scores for a set of events from the Hedge Fund dataset. It illustrates how events are distributed over time and also how specific events are predicted by the methods. Events are marked as vertical (red) lines. Moreover, scores are (min-max) scaled versions of the event (yes) class score (i.e. $Y_{t,1}$, as defined in Section \ref{sec::classifier_loss}).

We show events and scores for part of the training window (in gray) and also the complete test window (in white) of a particular fold. We also show the value of the S\&P500 index and some headlines of the day from the business section of the CNBC News website.\footnote{\url{https://www.cnbc.com}} These visualizations provide important insights regarding the nature of events in these datasets and also about the performance of event detection schemes.

Notice that prediction scores are correlated with events across methods, particularly the top-performing ones. However, most methods tend to suffer from false positive errors. This shows that the relationship between the graph dynamics and events is often weak (or noisy), which is a challenge for event detection methods.

Prediction scores also give us a better understanding of the differences between our method (\dyged) and baselines EvolveGCN and ASTGCN. Different from \dyged, these baselines capture the micro (or node-level) dynamics of the graph. On the other hand, our approach focuses on the macro (or graph-level) dynamics. First, \dyged is effective at fitting the training data. Moreover, \dyged also achieves better performance during testing, producing significantly fewer false positives. These results are consistent with the quantitative analysis provided in Section \ref{sec::results_accuracy}.

\begin{figure*}[ht]
\centering
\subfloat[\dyged Attention Weights]{
	\includegraphics[width=.24\textwidth]{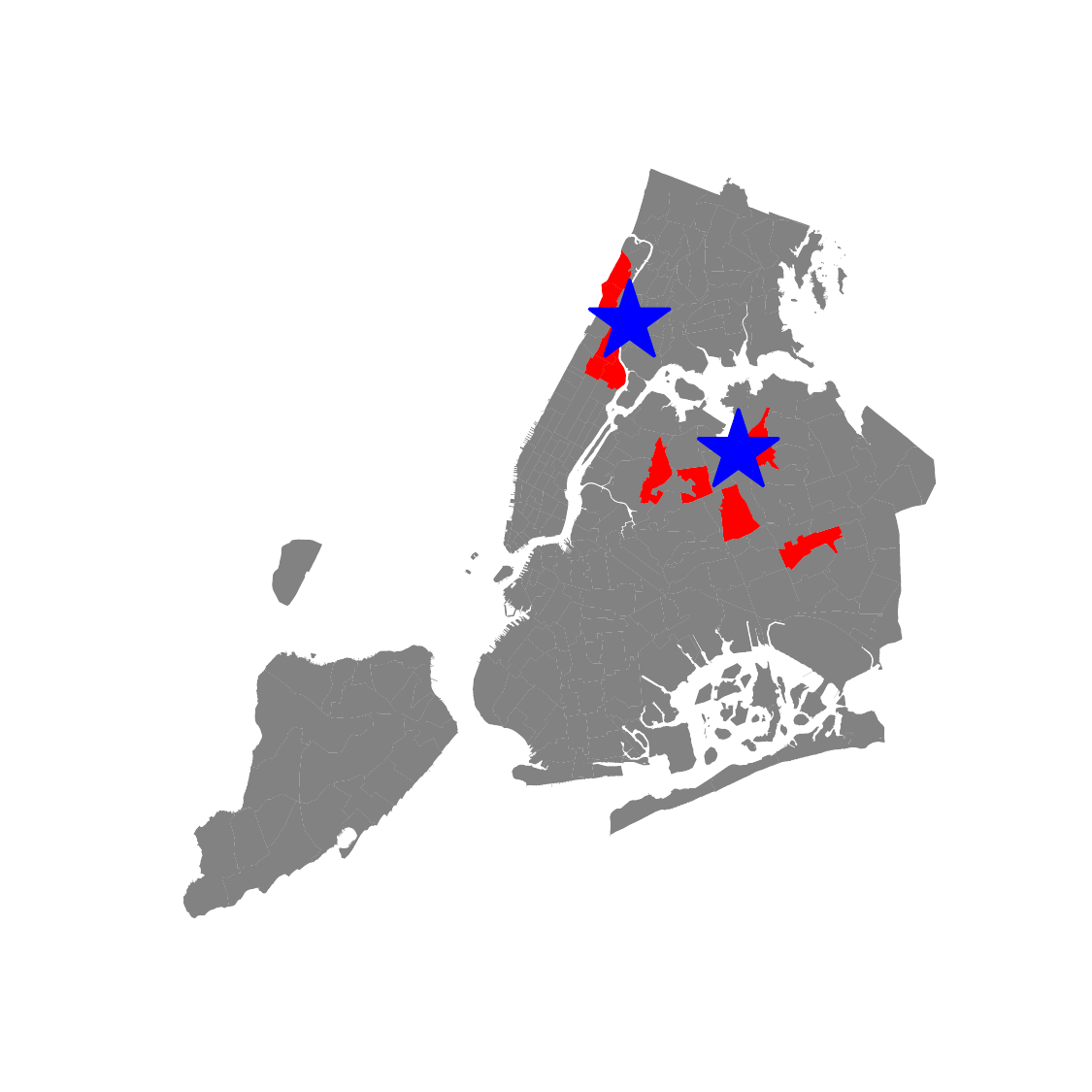}}
	\subfloat[Betweenness]{
	\includegraphics[width=.24\textwidth]{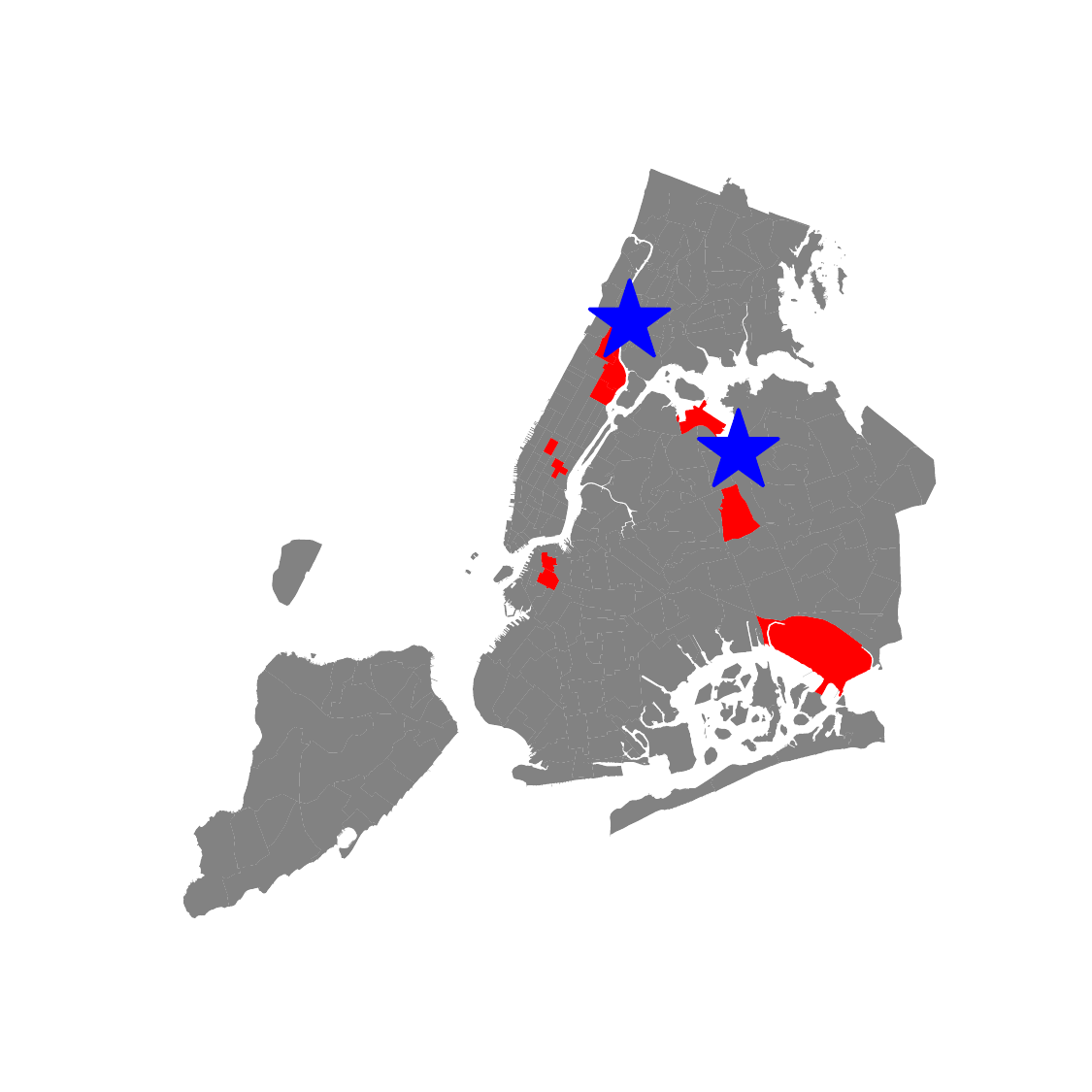}}
	\subfloat[Clustering]{
	\includegraphics[width=.24\textwidth]{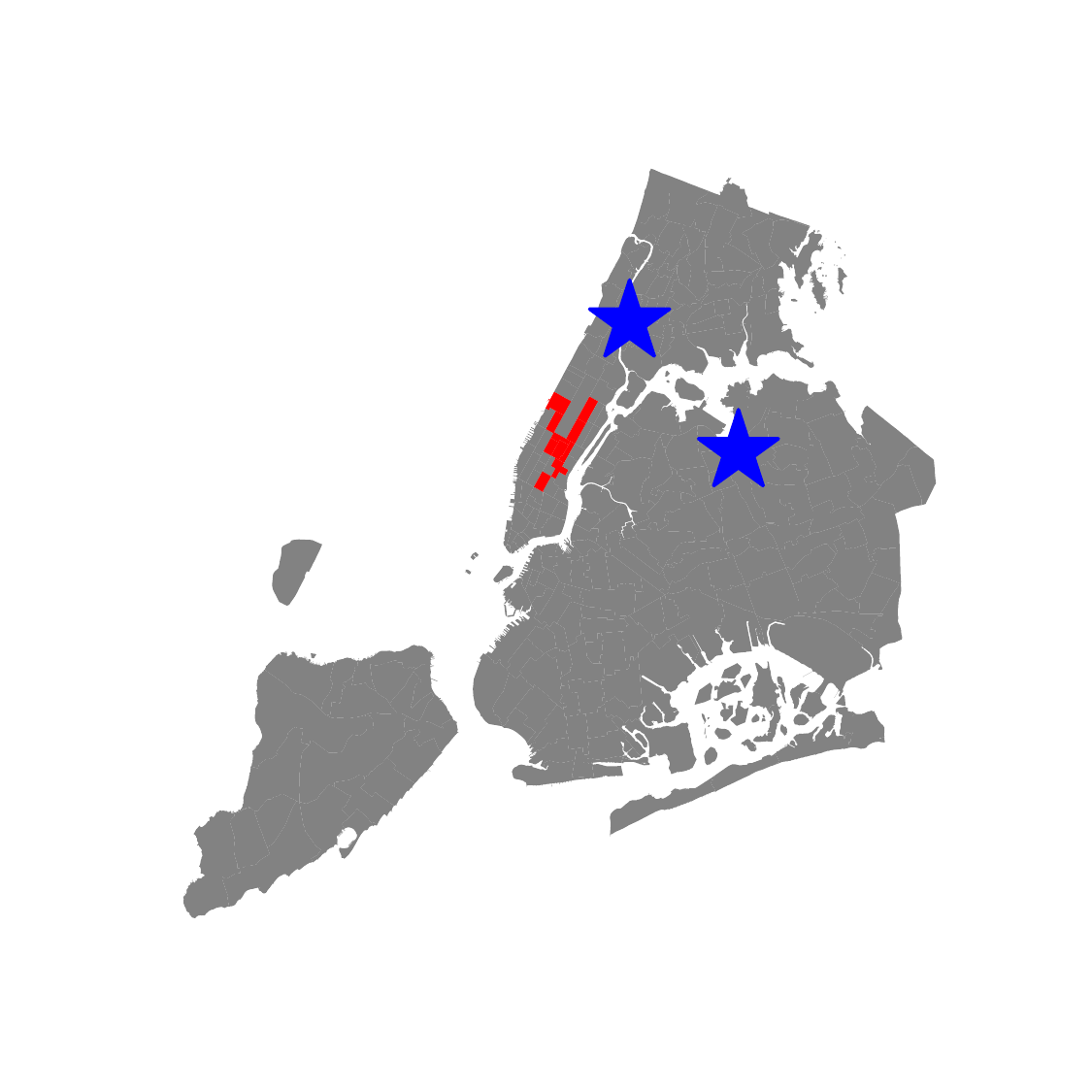}}
    \subfloat[Degree]{
	\includegraphics[width=.24\textwidth]{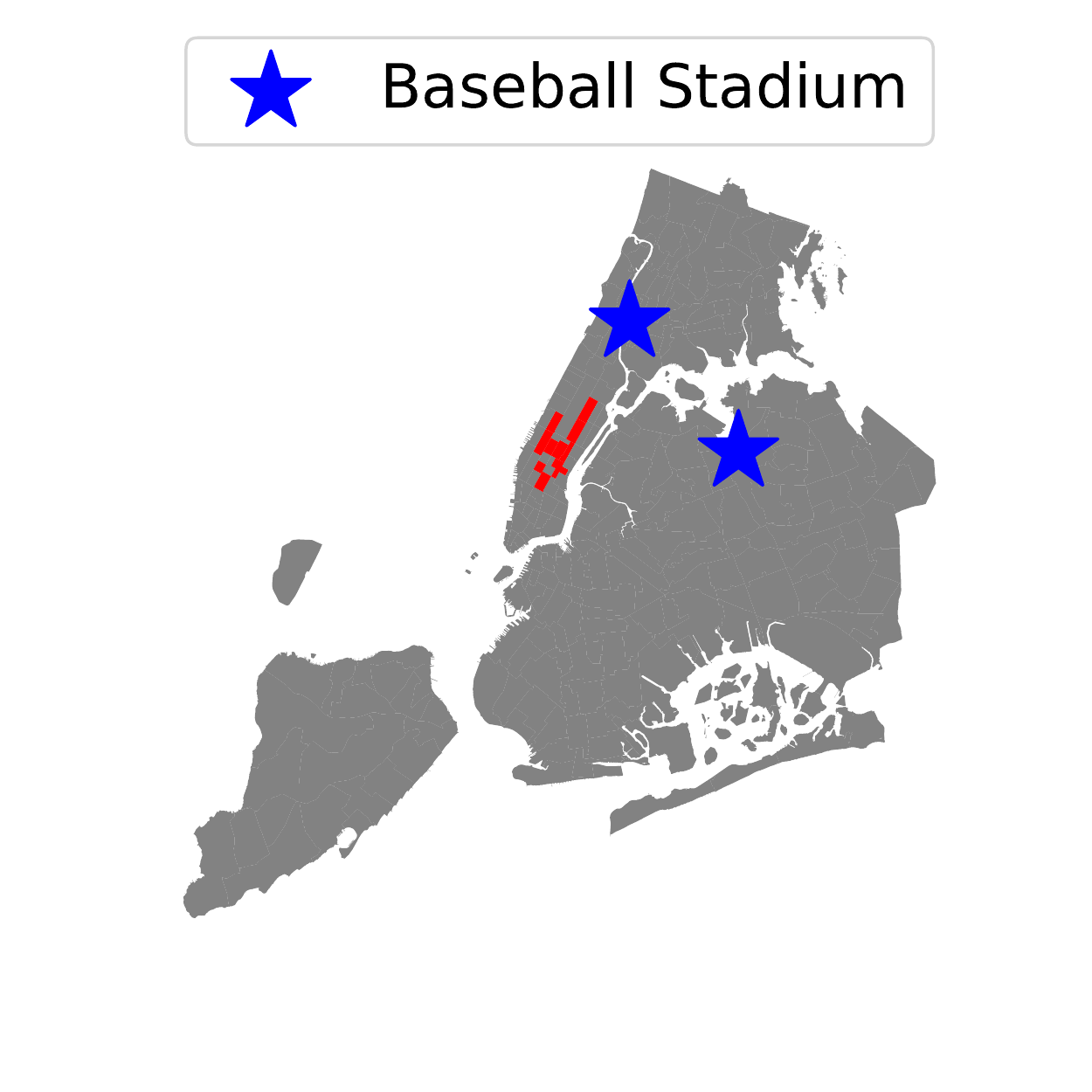}} 
	\caption{Top 10 taxi zones (in red) based on various metrics for NYC Cab dataset. Attention mechanism finds closer taxi zones to the stadiums compared to some node statistics such as betweenness centrality, clustering coefficient, and node degree. In other words, \dyged detects more crucial nodes for Baseball game detection.} %
\label{fig::taxi_zones_node_importance}
\end{figure*}

\begin{figure}[htbp]
\centering
    \subfloat[NYC Cab]{
	\includegraphics[width=.16\textwidth]{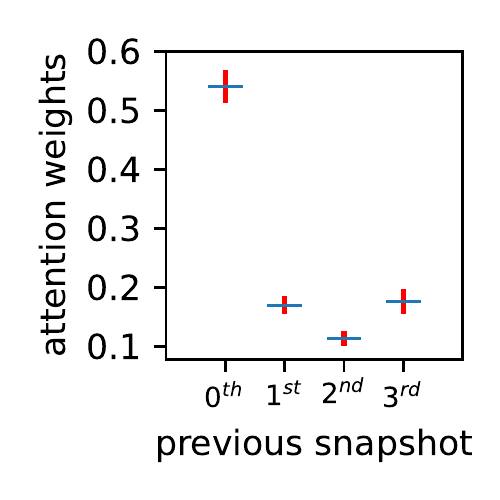}}
    \subfloat[Hedge Fund]{
	\includegraphics[width=.16\textwidth]{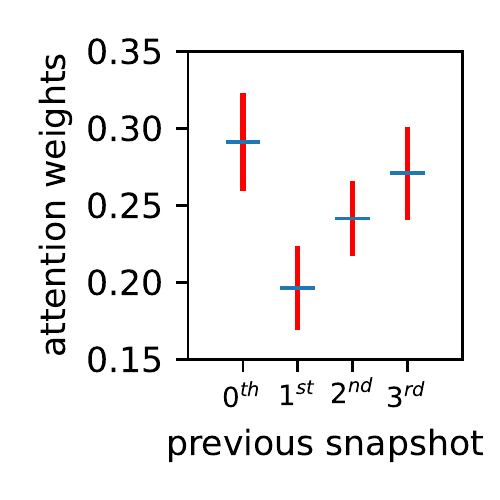}}
	\subfloat[Twitter Weather]{
	\includegraphics[width=.16\textwidth]{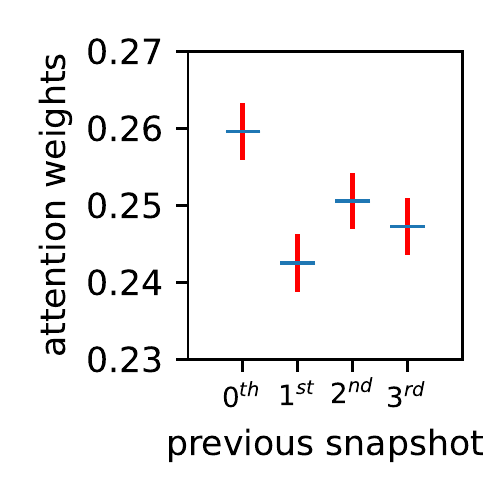}}
	\caption{Illustration of the attention weights (normalized) of current and previous three snapshots for all datasets. While in NYC Cab, the current snapshot has more attention weights than the previous ones, attention weights in the Hedge Fund and Twitter datasets reveal more about the importance of temporal information. Results show that history plays a role in predicting the event. $0^{th}$ refers to the current snapshot.}
\label{fig::time_attention}

\end{figure}

\subsection{Importance via Attention}
\label{sec::attention_results}

\dyged applies node and time self-attention weights for event detection. Here, we analyze these attention weights as a proxy to infer \textit{importance} node and time importance. We notice that the use of attention weights for interpretability is a contentious topic in the literature \cite{jain2019attention, pruthi2019learning,wiegreffe2019attention}. Still, we find that these learned weights to be meaningful for our datasets. They also provide interesting insights regarding the role played by self-attention in our model.

\subsubsection{Node Attention} A critical task in event detection on graphs is to measure the importance of nodes and subgraphs \cite{ying2019gnnexplainer} based on the events of interest. As a step towards answering this question, we analyze attention weights learned the v-Att(.) operator---normalized by the softmax function. For each node, we compute the average weight learned. For a comparison, we also consider the following classical node importance measures from the network science literature: \textit{degree, betweenness centrality}, and \textit{clustering coefficient}.

Figure \ref{fig::taxi_zones_node_importance} shows the top 10 most important taxi zones based on each importance measure for the NYC Cab dataset. Our attention weights find taxi zones near the baseball stadiums, whereas topology-based baseline measures select stations in downtown Manhattan and the airports. In Twitter Weather, where nodes are words, our solution set contains ``fire'', ``warm'', ``tree'', and ``snow'' as the top words while the topology-based baselines have ``weather'', ``update'', ``fire'', and ``barometer''. The results show that words found using our measure are more strongly associated with events of interest.

\subsubsection{Time (Snapshot) Attention} We also propose a time importance module that uses temporal self-attention weights via the function t-Att(.), to measure how the past snapshots (time) affect event detection. Similar to the node attention weights, the attention values here also signify the value of the information from the snapshots. We use three previous (i.e., $k=4$) and the current snapshot in our experiments. Figure \ref{fig::time_attention} shows the attention weights (output of softmax) for snapshots (with mean and standard deviation). For NYC Cab, the current snapshot has significantly higher weights. However, the remaining datasets reveal more interesting attention patterns. For instance, in Hedge Fund, the importance of earlier weights can be associated with the definition of an event---a stock market shock. For Twitter Weather, events often last several days, and thus weights are expected to be more uniformly distributed.

\begin{figure*}[ht]
\centering
\subfloat[NYC Cab]{
	\includegraphics[width=.32\textwidth]{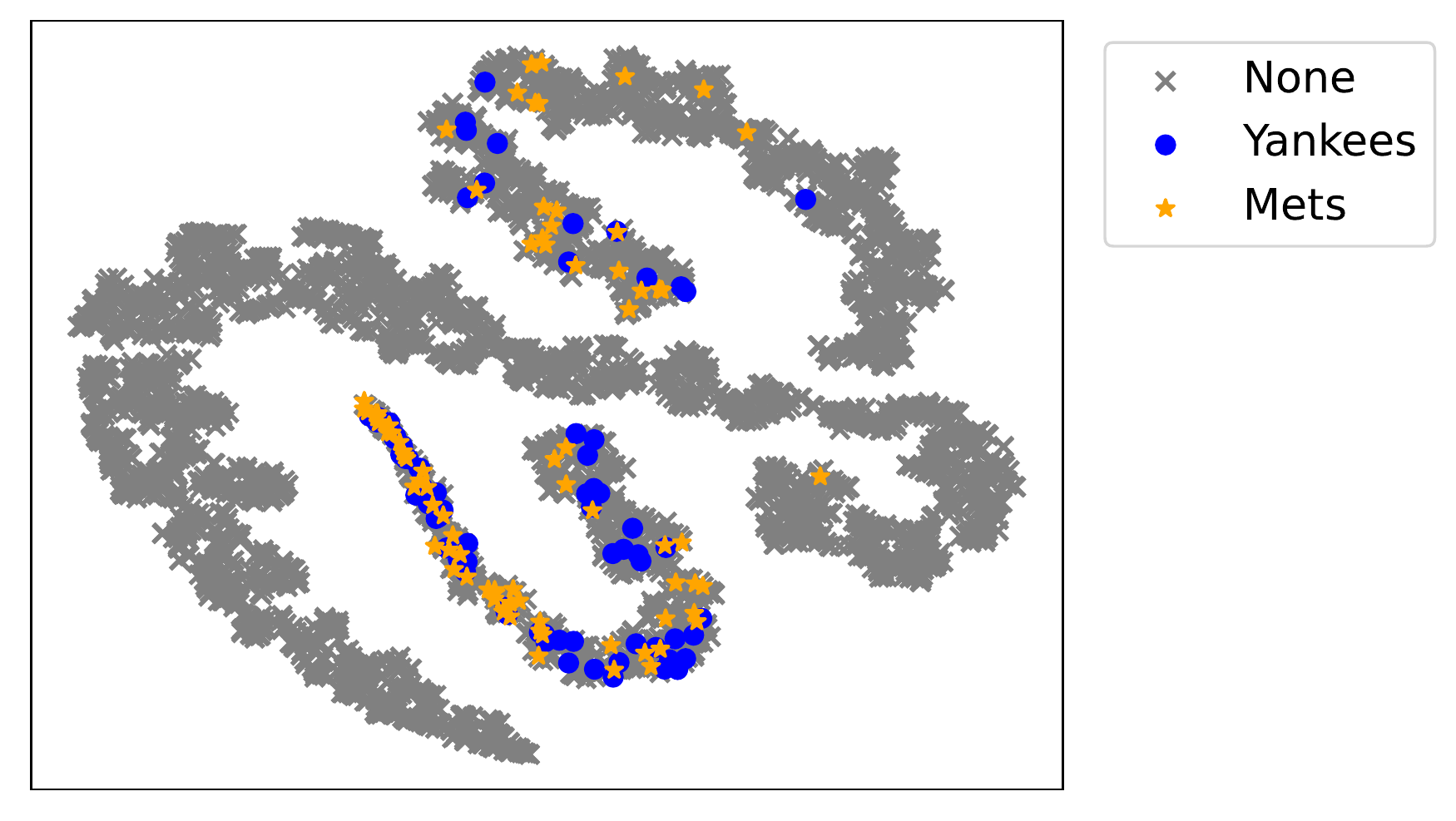}}
\subfloat[Hedge Fund]{
	\includegraphics[width=.3\textwidth]{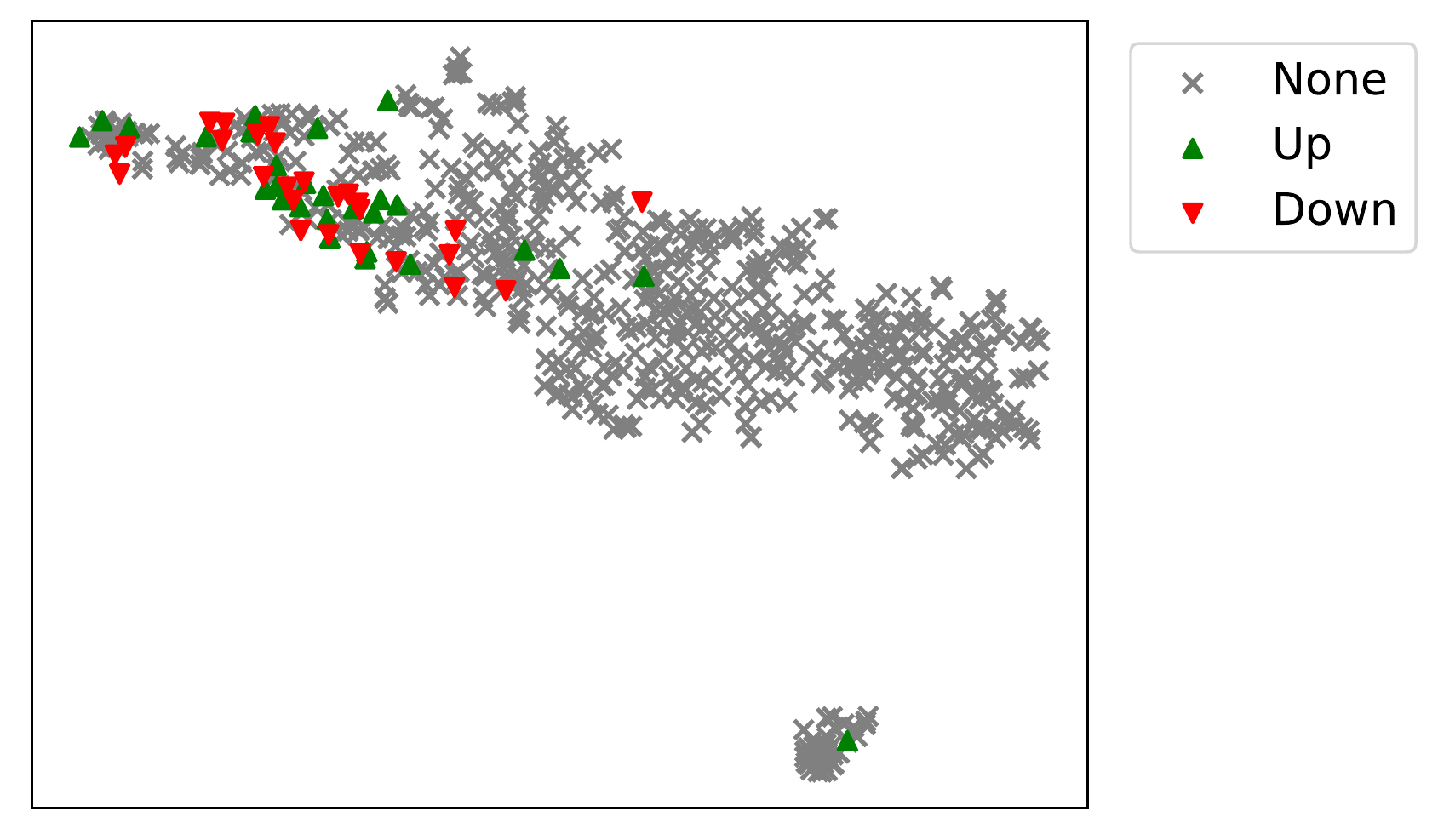}} 
	\subfloat[Twitter Weather]{
	\includegraphics[width=.32\textwidth]{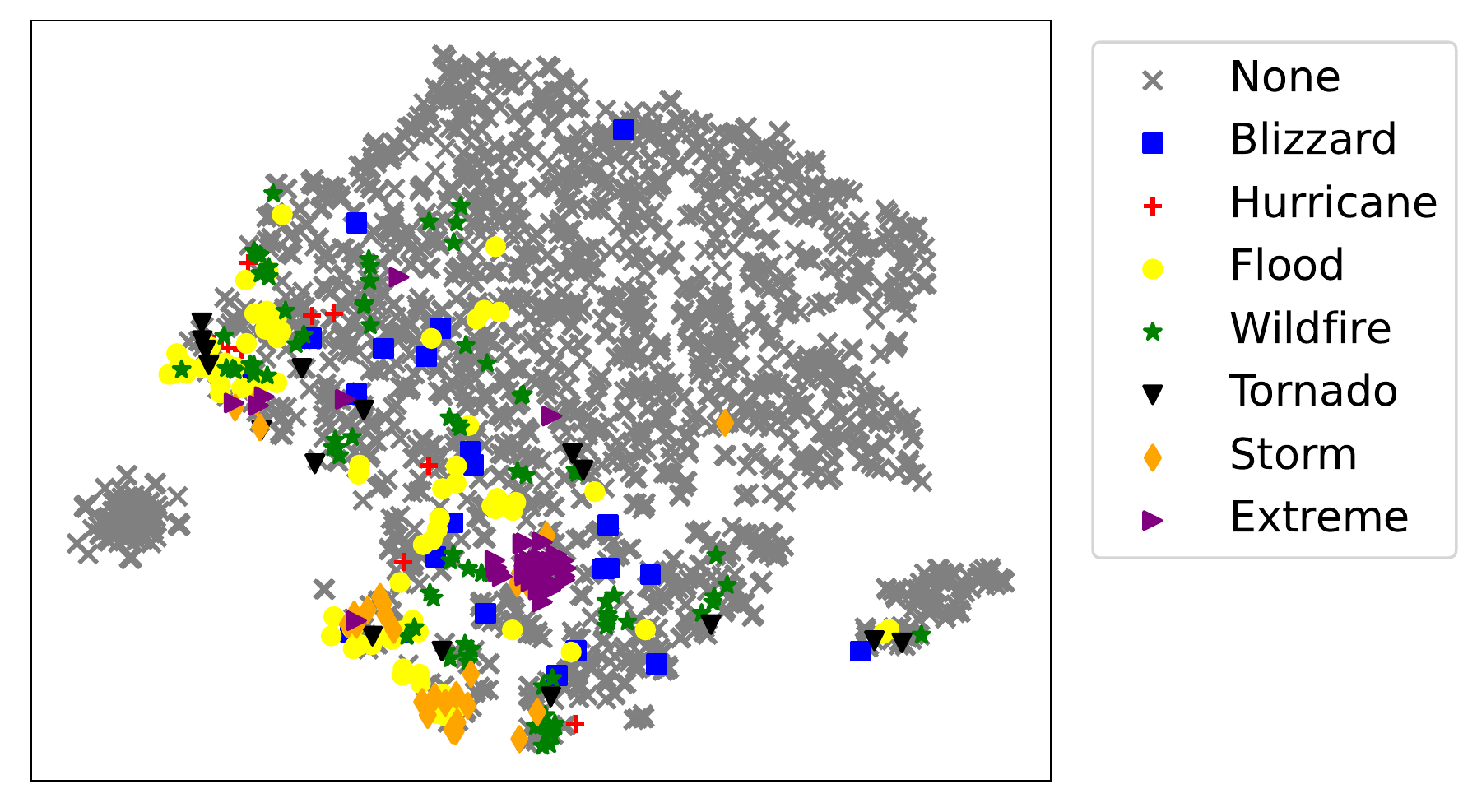}}
	
	\caption{An illustration of how DyGED graph embeddings support event diagnosis for three datasets. For each snapshot window $G_{t-k:t}$, we assign an event case (e.g. game at Yankees or Mets stadium for NYC Cab) or None. Results show that event causes tend to form clusters in the embedding for all datasets. Thus, events can be automatically diagnosed based on a few manually diagnosed ones in their embedding neighborhood. Better seen in color.} 
\label{fig::diagnosis}

\end{figure*}

\subsection{Event Embeddings and Diagnosis}
\label{sec::event_diagnosis}

Figure \ref{fig::diagnosis} shows graph embeddings produced by \dyged for our three datasets. Each point corresponds to a sequence of snapshots $G_{t-k:t}$. These are the same embeddings given as input to the MLP in our architecture. We set the number of dimensions $h$ of the embeddings originally to 64 and then project them to 2-D using tSNE \cite{van2008visualizing}. We also annotate each embedding with the type/cause of the event. For NYC Cab, we consider the stadium (\textit{Yankees} or {Mets}) where the game takes place. For Hedge Fund, we distinguish between \textit{up} and \textit{down} shocks. For Twitter Weather, we use the classification from the US National Weather Service (\textit{blizzard}, \textit{hurricane}, \textit{flood}, \textit{wildfire}, \textit{tornado}, \textit{storm} and \textit{extreme}).\footnote{\url{https://www.ncdc.noaa.gov/stormevents/pd01016005curr.pdf}} Notice that this additional information is not used for event detection.

As desired, events often form clusters in the embedding space for all datasets. That illustrates the ability of \dyged to produce discriminative representations that enable the classifier to identify events. However, notice that events and non-events are not completely separated in 2-D embeddings produced by \dyged which outperform all the baselines. This suggests that the event detection problem on dynamic graphs might be non-trivial.

It is also interesting to analyze whether \dyged embeddings can be useful for \textit{event diagnosis}, which consists of providing users with information that enables the identification of causes for the events \cite{julisch2003clustering,jeron2006supervision}. More specifically, embeddings might capture not only whether an event occurs or not but also the nature of the event---i.e. events with the same cause are embedded near each other. Our results show that events of the same type/cause tend to cluster in the embedding space. In particular, in the Twitter Weather dataset, events of type `Extreme' and `Storm' could be potentially diagnosed using a simpler nearest neighbors scheme.

\subsection{Event Detection Efficiency}
\label{sec::dyged_efficiency}

Table \ref{tab:test_times} shows testing times (in secs.) for a batch size of $100$ data points for NYC Cab, Hedge Fund, Twitter Weather, and Twitter Weather Large, respectively. The window size is set to 4. Results show that \dyged is scalable---up to 206 and 29 times faster than the top 2 baselines (ASTGCN and EvolveGCN, respectively).

\begin{table}[h]
\small
\centering
\setlength{\tabcolsep}{4pt}
\renewcommand{\arraystretch}{1}
\begin{tabular}{ccccc}%
\toprule
& \textbf{NYC Cab} & \textbf{Hedge Fund} & \textbf{TW} & \textbf{TW-Large} \\
\midrule
\textit{EvolveGCN} & 1.912 & 0.303 & 0.949 & 4.65 \\
\textit{ASTGCN} & 13.57 & 2.131 & 6.916 & 24.41 \\
\textit{DynGCN} & 0.064 & 0.019 & 0.080 & 0.622  \\
\textit{DiffPool} & 0.057 & 0.016 & 0.045 & 0.482 \\
\textit{\dyged} & 0.066 & 0.017 & 0.048 & 0.479 \\
\bottomrule
\end{tabular}
\caption{Testing times (in secs.) for all the methods with windows length \textit{k+1} set to 4. \dyged is scalable and up to 206 times faster than the best baseline, ASTGCN.}
\label{tab:test_times}
\end{table} 

\subsection{Ablation Study}
\label{sec::ablation_study}

\subsubsection{Feature Variation}
\label{sec::dynamic_node_features}

We also calculate three graph statistics as dynamic node features: node degree, betweenness centrality, and clustering coefficient. Table \ref{tab:dynamic_static} shows results for static-only, dynamic-only, and static+dynamic node features for \dyged and the best baseline (ASTGCN) using all datasets. Results show that static features are often more relevant than dynamic ones---NYC Cab is the exception. Moreover, combining static and dynamic features often improves performance. We notice that for Twitter Weather (TW), static features (word embeddings) are quite expressive.

\begin{table}[ht]
\centering
\footnotesize
\setlength{\tabcolsep}{4pt}
\renewcommand{\arraystretch}{1}
\begin{tabular}{cccc}%
\toprule
\textbf{Method} & \textbf{NYC Cab}& \textbf{Hedge Fund} & \textbf{Twitter Weather} \\
\midrule
ASTGCN-S & $0.903$ & $0.753$ & $0.774$ \\
\dyged-S & $0.905$ & $\underline{0.787}$ & $\underline{0.810}$ \\
ASTGCN-D & $0.894$ & $0.741$ & $0.747$ \\
\dyged-D & $\underline{0.911}$ & $0.786$ & $0.765$ \\
ASTGCN-SD & $0.905$ & $0.763$ & $0.751$ \\
\dyged-SD & $\textbf{0.925}$ & $\textbf{0.798}$ & $\textbf{0.815}$ \\
\bottomrule
\end{tabular}
\caption{AUC scores of \dyged and the best baseline method ASTGCN with different sets of node features. \textit{-S}, \textit{-D}, \textit{-SD} suffixes are for static-only, dynamic-only, static+dynamic node features, respectively. TW is the Twitter Weather dataset. \label{tab:dynamic_static}}
\end{table}

\subsubsection{Pooling Operator Variation}
\label{sec::pooling_operator_varitaion}

Mean and max operators are two of the most common pooling operators in convolutional networks \cite{goodfellow2016deep}. They have been used for down-sampling the data and their usefulness depends on the application and the dataset \cite{boureau2010theoretical}. We compare these operators against self-attention graph pooling (Section \ref{sec::pooling}). Table \ref{tab:pooling_methods} shows that the self-attention operator (\dyged) outperforms the other operators for all datasets. These results highlight the importance of adaptive node weighting when capturing the macro dynamics for event detection, one of the essentials of our approach.

\begin{table}[ht]
\centering
\footnotesize
\setlength{\tabcolsep}{4pt}
\renewcommand{\arraystretch}{1}
\begin{tabular}{cccc}%
\toprule
\textbf{Method} & \textbf{NYC Cab}& \textbf{Hedge Fund} & \textbf{Twitter Weather} \\
\bottomrule
\dyged-Mean & $\underline{0.879}$ & $0.704$ & $\underline{0.773}$ \\
\dyged-Max & $\underline{0.880}$ & $\underline{0.729}$ & $0.729$ \\
\dyged & $\textbf{0.905}$ & $\textbf{0.787}$ & $\textbf{0.810}$ \\
\bottomrule
\end{tabular}
\caption{AUC scores of  our event detection method \dyged equipped with different pooling operators. Attention pooling outperforms mean and max pooling for all datasets using our method \dyged. TW is the Twitter Weather data.
\label{tab:pooling_methods}}
\end{table}

\section{Conclusions}
This paper is focused on event detection on dynamic graphs. We have proposed a deep learning based method, \dyged, which learns correlations between the graph macro dynamics---i.e. a sequence of temporal graph representations---and events. We compared \dyged against multiple baselines using a representative set of datasets. Our approach outperformed the baselines in accuracy while being more scalable than the most effective one. We also showed how our method could be applied for event diagnosis as well as to provide interpretability via self-attention on nodes and snapshots. In future work, we want to develop hierarchical event detection architectures that are able to combine macro and micro dynamics. We are also interested in designing even more interpretable models for event detection on graphs so that the discovered events can be associated with subgraphs and their dynamics.

\section{Acknowledgments}

This work is funded by NSF via grant IIS 1817046. Some material is based upon work supported by the Air Force Office of Scientific Research under Minerva award number FA9550-19-1-0354.

\bibliography{aaai23}

\end{document}